\documentclass[10pt,twocolumn,letterpaper]{article}

\usepackage{cvpr}              

\usepackage{graphicx}
\usepackage{amsmath}
\usepackage{amssymb}
\usepackage{booktabs}
\usepackage{float}
\usepackage{makecell}
\usepackage[table]{xcolor}
\usepackage{arydshln}
\usepackage[accsupp]{axessibility}

\usepackage[pagebackref,breaklinks,colorlinks]{hyperref}

\usepackage[capitalize]{cleveref}
\crefname{section}{Sec.}{Secs.}
\Crefname{section}{Section}{Sections}
\Crefname{table}{Table}{Tables}
\crefname{table}{Tab.}{Tabs.}

\def\cvprPaperID{7} 
\def\confName{CVPR}
\def\confYear{2023}

\begin{document}


\title{EGA-Depth: Efficient Guided Attention for Self-Supervised \\ Multi-Camera Depth Estimation}

\author{
Yunxiao Shi$^{1}$~~~
Hong Cai$^{1}$~~~
Amin Ansari$^{2}$~~~
Fatih Porikli$^{1}$~~~
\smallskip
\\
$^{1}$Qualcomm AI Research$^{\dagger}$\quad $^{2}$Qualcomm Technologies, Inc.
\\
\smallskip
{\tt\small\{yunxshi, hongcai, amina, fporikli\}@qti.qualcomm.com}
}

\maketitle

\begin{abstract}
\vspace{-7pt}
The ubiquitous multi-camera setup on modern autonomous vehicles provides an opportunity to construct surround-view depth. Existing methods, however, either perform independent monocular depth estimations on each camera or rely on computationally heavy self attention mechanisms. 
In this paper, we propose a novel guided attention architecture, EGA-Depth, which can improve both the efficiency and accuracy of self-supervised multi-camera depth estimation. More specifically, for each camera, we use its perspective view as the query to cross-reference its neighboring views to derive informative features for this camera view. This allows the model to perform attention only across views with considerable overlaps and avoid the costly computations of standard self-attention. Given its efficiency, EGA-Depth enables us to exploit higher-resolution visual features, leading to improved accuracy. Furthermore, EGA-Depth can incorporate more frames from previous time steps as it scales linearly w.r.t. the number of views and frames.
Extensive experiments on two challenging autonomous driving benchmarks nuScenes and DDAD demonstrate the efficacy of our proposed EGA-Depth and show that it achieves the new state-of-the-art in self-supervised multi-camera depth estimation.

\end{abstract}

\begin{figure}[t!]
    \vspace{-12pt}
    \centering
    \includegraphics[width=0.5\textwidth]{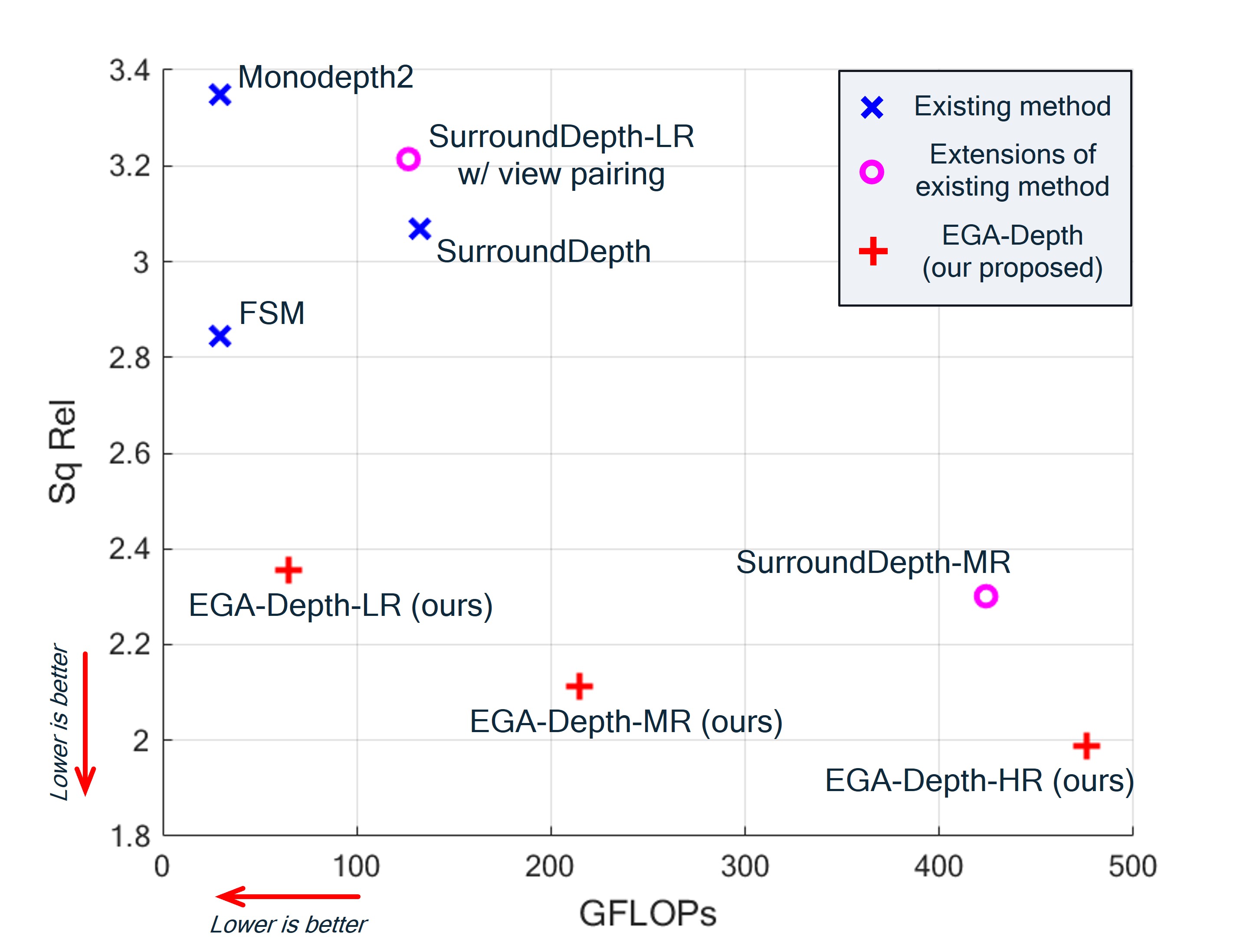}
    \vspace{-20pt}
    \caption{Accuracy (Squared Relative Error) vs. efficiency (GFLOPs). Our proposed EGA-Depth achieves the best accuracy-efficiency trade-off when comparing to baseline and latest state-of-the-art methods, including Monodepth2~\cite{godard2019digging}, Full Surround Monodepth (FSM)~\cite{guizilini2022full}, and SurroundDepth~\cite{wei2022surrounddepth}. We also compare with extensions of SurroundDepth (implemented by us), e.g., using only pairs of views for self-attention, increasing feature map resolutions by replacing standard self-attention with Linformer~\cite{wang2020linformer}. Both of these designs under-perform our EGA-Depth. LR, MR, and HR indicate the low, median, and high resolution feature maps fed to the attention module.}
    \label{fig:accuracy-efficiency-tradeoff}
    \vspace{-10pt}
\end{figure}

\vspace{-5pt}
\section{Introduction}
\label{sec:intro} \vspace{-3pt}

{\let\thefootnote\relax\footnotetext{{
\hspace{-6.5mm} $\dagger$ Qualcomm AI Research is an initiative of Qualcomm Technologies, Inc.}}}

Depth plays a fundamental role in 3D perception, which is key to various applications including autonomous driving, AR/VR, and robotics. While it is possible to measure depth using LiDAR or Time-of-Flight (ToF) sensors, such specialized hardware can be expensive, consume a lot of power, require high-fidelity cross-sensor calibration, and fail to obtain depth for certain surfaces. On the other hand, inferring depth from camera images is more cost-efficient and can still provide promising accuracy. Traditional approaches~\cite{saxena2007depth, furukawa2010towards, newcombe2011dtam} utilize stereoscopic vision and/or structure-from-motion to estimate depth. However, these methods have limited accuracy. By leveraging deep learning, researchers have been able to achieve significantly more accurate depth estimation~\cite{eigen2014depth,fu2018deep,zhu1232020mda,bhat2021adabins,ranftl2021vision,yuan2022neural}.

Learning-based methods, however, require a massive amount of densely labeled high-quality ground-truth depth maps for training, which is costly if not impractical, to acquire at scale using range-finding sensors. To overcome this challenge, self-supervised learning \cite{zhou2017unsupervised,monodepth17,godard2019digging} has emerged as a new paradigm to train depth estimation networks without the need for ground truth depths. Subsequent works have looked at various aspects to improve self-supervised depth estimation in terms of more advanced architectures~\cite{watson2021temporal, guizilini2022multi, zhao2022monovit} and better training procedures~\cite{guizilini2022learning, cai2021x}. However these works only focus learning using a single camera.

More related to our work are those that consider multi-camera depth estimation, which allows one to potentially attain a 360$^{\circ}$ view of the surrounding environment. Guizilini~\textit{et al.}~\cite{guizilini2022full} incorporates spatial-temporal view relationships during training but still processes each camera view separately at test time. More recently, Wei~\textit{et al.}~\cite{wei2022surrounddepth} proposed a transformer-based architecture to jointly process multiple camera views with standard self-attention. While this method can exploit cross-view information at test time, the expensive self-attention limits the resolution of visual features that can be processed, leading to sub-optimal results.

In this paper, we propose a novel Efficient Guided Attention architecture, EGA-Depth, in order to address the existing shortcomings and improve both the efficiency and accuracy of self-supervised multi-camera depth estimation. We first develop a guided attention mechanism that for each camera view, allows interaction between the neighboring view features. Specifically, when processing the features for each camera, we obtain queries from the current view, keys and values from the stacked features of the neighboring views. These are then fed into an efficient attention module. This allows our proposed model to achieve similar or better accuracy compared to the SOTA while using considerably less computation as can be seen in Figure~\ref{fig:accuracy-efficiency-tradeoff}.


Our proposed guided attention makes it possible to efficiently scale up the model complexity in order to boost depth estimation accuracy. More specifically, we can exploit higher resolutions for the visual features which allows the model to utilize more visual details for depth estimation. Furthermore, EGA-Depth can incorporate more frames from previous time steps, which allows to leverage temporal visual correlations to further improve accuracy.


Our main contributions are summarized as follows:
\begin{itemize}
    \vspace{-3pt}
    \item We propose an efficient guided attention scheme for self-supervised multi-camera depth estimation models, where each camera view cross-references its neighboring views.     
    Our design can significantly reduce computation costs while maintaining accuracy.
    \vspace{-3pt}
    \item Based on our efficient guided attention, we can efficiently exploit higher-resolution visual features as well as leverage views from previous time steps, which improves depth estimation accuracy.
    \vspace{-3pt}
    \item Our method acheives state-of-the-art results with optimal accuracy-efficiency trade-off on two large-scale self-supervised multi-camera depth estimation benchmarks, i.e. nuScenes \cite{caesar2020nuscenes} and DDAD \cite{packnet}. 
\end{itemize}


\section{Related Works}
\label{sec:relatedworks}
\vspace{-5pt}



\begin{figure*}[t!]
    \centering
    \includegraphics[width=0.9\textwidth]{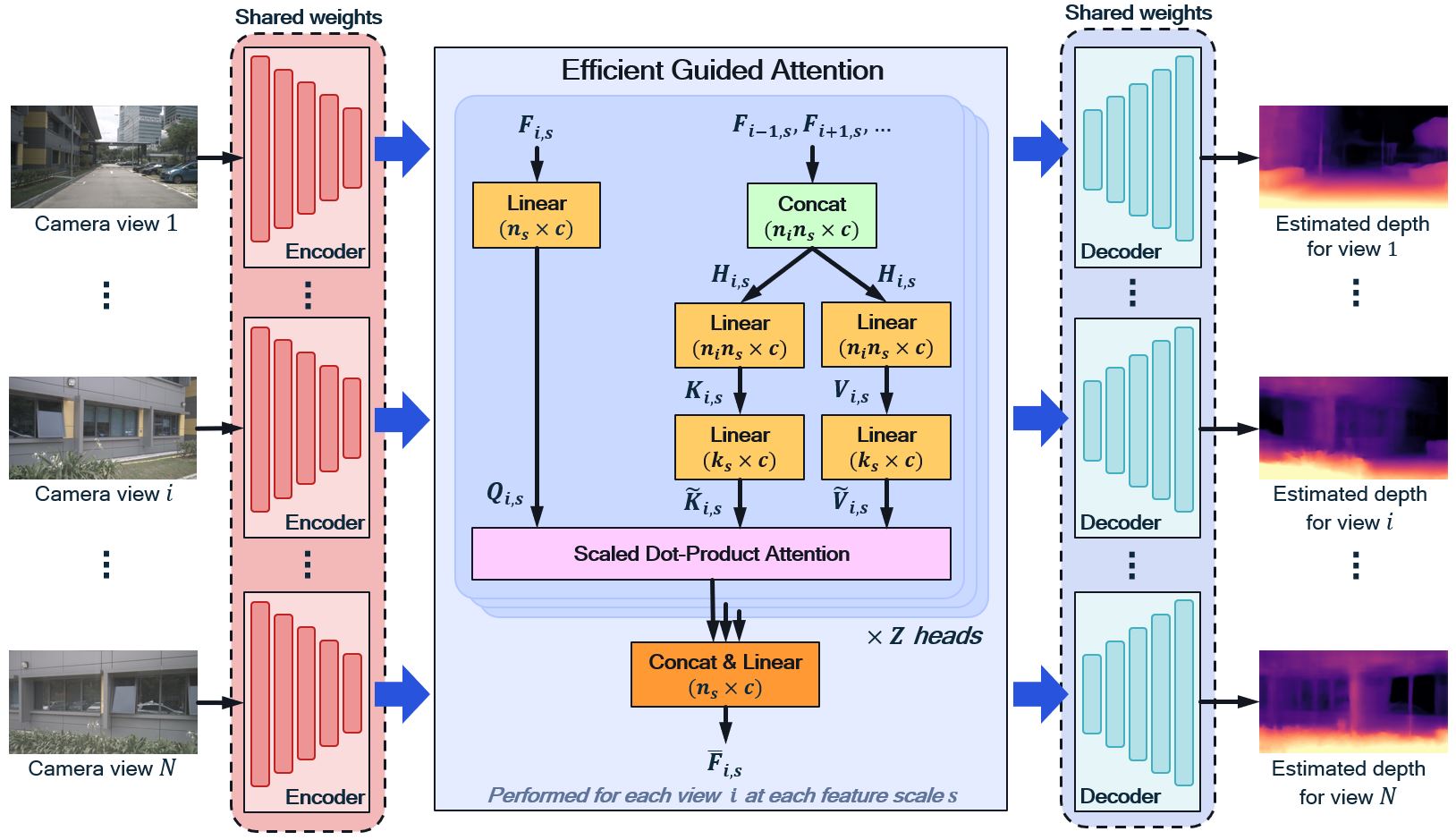}
    \vspace{-10pt}
    \caption{Overview of our proposed EGA-Depth attention architecture. A single ResNet34 encoder~\cite{he2016deep} first extracts features from the input multi-camera images through weight sharing. The extracted features are then fed to our efficient guided attention module. Standard norm layers and skip connections are used in the attention module, which are not shown in the diagram for a more concise illustration. Lastly the attended features are consumed by a single decoder through shared weights to output the final estimated depth maps.}
    \label{fig:system}
    \vspace{-8pt}
\end{figure*}

\textbf{Self-Supervised Depth Estimation:} Zhou~\textit{et al.}~\cite{zhou2017unsupervised} pioneered the use of self-supervised learning for depth estimation by casting the problem as differentiable view synthesis, which employs two networks to predict depth and camera pose while masking out regions that violate the static scene assumption. Godard~\textit{et al.}~\cite{godard2019digging} improved the results by proposing the per-pixel minimum re-projection loss, full-resolution multi-scale sampling, and auto-masking the pixels where no camera motion is observed. Later works seek further improvements in various aspects, e.g., improving the photometric matching~\cite{shi2019self,shu2020feature, jiang2020dipe}, handling dynamic objects~\cite{casser2019depth, gordon2019depth, dai2020self, klingner2020self}, utilizing semantic segmentation~\cite{tosi2020distilled, guizilini2020semantically, cai2021x}, supplying temporal information to the network~\cite{patil2020don, watson2021temporal, fan2021multiscale}, etc. Recently, Guizilini~\textit{et al.}~\cite{guizilini2022full} extended self-supervised monocular depth estimation to the full surrounding multi-camera setting by introducing spatial-temporal contexts and pose consistency constraints in training. Wei~\textit{et al.}~\cite{wei2022surrounddepth} proposed a cross-view transformer to capture interactions between cameras for multi-camera depth estimation. 

\vspace{2pt}
\textbf{Vision Transformers and Self-Attention:}
Inspired by the astounding success of transformers~\cite{vaswani2017attention,devlin2018bert,brown2020language} in Natural Language Processing (NLP), the interest in exploring transformers for vision tasks~\cite{dosovitskiy2020image,Liu_2021_ICCV,touvron2021training,fan2021multiscale,li2022mvitv2,carion2020end} has been soaring. At the core of transformers is the Multi-Headed Self-Attention (MHSA) mechanism~\cite{vaswani2017attention} that computes a self-attention matrix, where the model exhaustively compares each token to every other token. This allows the model to extensively identify and exploit correlations within the input. 
This, however, has a quadratic complexity w.r.t. to the token sequence length that becomes a computation bottleneck. For computer vision tasks, this limitation is more pronounced since even an image of moderate resolution will result in a very long sequence length. 

\vspace{2pt}
\textbf{Linear Transformers:}
Recently, various techniques have been proposed to reduce the computation complexity of self-attention from quadratic to linear. Wang~\textit{et al.}~\cite{wang2020linformer} propose the projection of the keys and values to a fixed lower dimension. Kitaev~\textit{et al.}~\cite{kitaev2020reformer} adopts a multi-round Locality Sensitive Hashing (LSH) strategy to select the most similar pairs and only computes self-attention between them. Different methods have been proposed to approximate the softmax-based self-attention in order to achieve linear complexity~\cite{choromanski2020rethinking, katharopoulos2020transformers, peng2021random, kasai2021finetuning, xiong2021nystromformer, lu2021soft, mehta2022separable}. We refer readers to recent surveys for more comprehensive coverage of linear and efficient transformer designs~\cite{yi2022efficient, khan2022transformers}.



\section{Method}
\vspace{-3pt}
Here we present our novel efficient guided attention architecture for self-supervised multi-camera depth estimation (EGA-Depth). We describe the detailed design of our guided attention scheme and analyze its efficiency in Section~\ref{subsec:EGA}. Section~\ref{subsec:high_res_temporal} describes how EGA-Depth enables using higher-resolution visual features as well as camera views from previous time steps, in an efficient manner, to enhance the depth estimation accuracy.


\subsection{Efficient Guided Attention}
\label{subsec:EGA} \vspace{-3pt}




We first provide an overview of the meta-architecture of our multi-camera depth estimation network. Consider a setup with $N$ cameras. For each camera $i\in\{1,...,N\}$, the captured images are fed into a single encoder to extract multi-scale visual feature maps, $F_{i,s}\in \mathbb{R}^{n_s \times c}$, where $s\in\{1,...,N_s\}$ denotes the feature map scale with $N_s$ being the number of scales,
$n_s=H_sW_s$ is the number of spatial elements in $F_{i,s}$, $H_s$ and $W_s$ denote the height and width of the feature maps at scale $s$, and $c$ denotes the number of feature channels. 
For notation simplicity, we assume that $F_{i,s}$ is flattened.
These feature maps are consumed by an attention module, which finds and utilizes their cross-correlations to refine the feature maps. We denote the output feature maps from the attention module as $\bar{F}_{i,s}$, for $i\in\{1,...,N\}$ and $s\in\{1,...,N_s\}$. For each camera, its updated multi-scale feature maps are then fed into a single decoder to generate the estimated depth maps.

When it comes to the attention, the existing SOTA model~\cite{wei2022surrounddepth} applies the standard self-attention to the feature maps of all the views at each scale. This incurs unnecessary computations and with the attention complexity being quadratic w.r.t. both the size of the feature map as well as the number of views, a large fraction of the computation is wasted on attending across views with little or no overlaps.

Our proposed efficient guided attention does not suffer from these limitations and provides an efficient alternative to the existing methods. First, for each $F_{i,s}$, we only utilize features of the neighboring views with considerable overlaps to compute attention. This allows our model to focus on views with meaningful overlaps and avoid spending computation over non-/little-overlapping views. Next, we replace standard self-attention with our guided attention. Specifically, $F_{i,s}$ is used to compute queries and the stacked features of the neighboring views, $H_{i,s} = \text{concat}(F_{i-1,s},F_{i+1,s},...)\in \mathbb{R}^{(n_i \cdot n_s)\times c}$, are used to compute keys and values, where $n_i$ is the number of neighboring views and $F_{i-1,s},F_{i+1,s},...$ denote their features. Formally, the queries, keys, and values are given by\vspace{-2pt}
\begin{equation}\label{eq:qkv_1}\vspace{-2pt}
Q_{i,s}=F_{i,s}W_{q,i,s},\text{ }K_{i,s}=H_{i,s}W_{k,i,s},\text{ }V_{i,s}=H_{i,s}W_{v,i,s}, 
\end{equation}
where $W_{q,i,s},\, W_{k,i,s},\, W_{v,i,s} \in \mathbb{R}^{c \times c}$ are learnable projection matrices. The attention is then calculated as follows:\vspace{-2pt}
\begin{equation}\label{eq:attention_1}\vspace{-2pt}
\text{softmax}\bigg(\underbrace{\frac{Q_{i,s}K_{i,s}^{\top}}{\sqrt{c}}}_\text{$n_s\times (n_i\cdot n_s)$}\bigg)\cdot\underbrace{\tilde{V}_i}_\text{$(n_i \cdot n_s)\times c$}
\end{equation}

It can be seen in Eq.~(\ref{eq:attention_1}) that when computing the attention map, our proposed guided attention only incurs a linear complexity w.r.t. the number of participating views. 
Moreover, as discussed, the number of participating views will be small as we only use those with meaningful overlaps. Although we require less computation, the critical functional ability to cross-reference different views is retained.

While the guided attention described above can already achieve better efficiency, its complexity is still quadratic w.r.t. the length of the features (i.e., $n_i\cdot n_s$). Inspired by the recent work of Linformer~\cite{wang2020linformer}, we perform further projections to bring the keys and values to a prescribed, input-invariant embedding dimension. Specifically,\vspace{-2pt}
\begin{equation}\vspace{-2pt}
\tilde{K}_{i,s} = P_{k,i,s}K_{i,s}, \quad \tilde{V}_{i,s} = P_{v,i,s}V_{i,s},
\end{equation}
where $P_{k,i,s},\, P_{v,i,s} \in \mathbb{R}^{k_s \times (n_i\cdot n_s)}$, $\tilde{K}_{i,s},\, \tilde{V}_{i,s} \in \mathbb{R}^{k_s \times c}$, and $k_s$ is a prescribed projection dimension for each feature scale. Using $Q_{i,s}$, $\tilde{K}_{i,s}$, and $\tilde{V}_{i,s}$, we have\vspace{-2pt}
\begin{equation}\vspace{-2pt}
\text{softmax}\bigg(\underbrace{\frac{Q_{i,s}\tilde{K}_{i,s}^{\top}}{\sqrt{c}}}_\text{$n_s\times k_s$}\bigg)\cdot\underbrace{\tilde{V}_{i,s}}_\text{$k_s\times c$}
\end{equation}
which scales linearly w.r.t. the size of input feature, $n_s$ (with an approximate choice of $k_s$ as shown in~\cite{wang2020linformer}).

Figure~\ref{fig:system} summarizes our proposed approach. Overall, it facilitates efficient attention across the multiple views, by focusing computation on meaningfully overlapping views and removing quadratic complexity from several aspects of the attention. Our efficient guided attention reaches new SOTA in accuracy with less computation, as we shall see in the experiments.

\begin{figure*}[t!]
  \centering
   \includegraphics[width=0.99\textwidth]{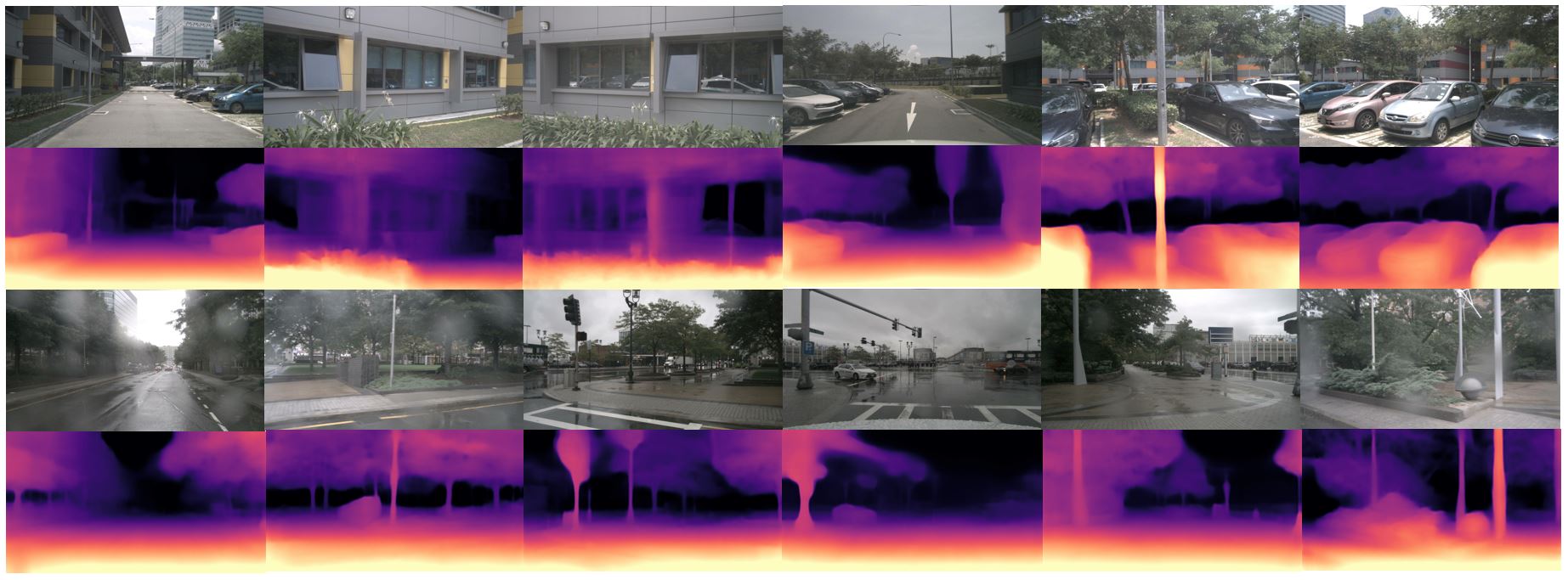}
   \vspace{-8pt}
   \caption{Qualitative results of self-supervised multi-camera depth predictions on two sample scenes from the nuScenes dataset.
   For each scene, we show the front, front-left, back-left, back, back-right, and front-right camera views from left to right. The first scene is under normal weather conditions and the second one is a rainy scene. We can see that even in adverse weather conditions, our EGA-Depth can generate accurate depth maps with fine details.}
   \label{fig:nuscenesres}
   \vspace{-5pt}
\end{figure*}

\subsection{Boosting Depth Estimation Accuracy}\label{subsec:high_res_temporal}\vspace{-3pt}
Given the elegant design of our efficient guided attention, it is computationally feasible to scale up the attention module. We can now utilize features of higher spatial resolutions and incorporate frames from the previous time steps. Application of these new venues enables us to further improve the depth estimation accuracy.

\textbf{High-Resolution Features:} In the existing SOTA method~\cite{wei2022surrounddepth}, due to the high complexity of self-attention, tiny spatial dimensions are used for the feature maps to be fed into the attention module in order to stay computationally feasible. For instance,  11$\times$20 feature maps are used for an input image of size 352$\times$640 at all feature scales. This severely limits the model's ability to perform accurate depth estimation. Given the quadratic complexity, attempts to directly increase the feature resolution would result in infeasible training and inference even with multiple modern high-end GPUs (\textit{e.g.}, 4 Nvidia A100s).

With our proposed EGA scheme, it is possible to use higher resolutions for the features, e.g., 4$\times$ larger, since the attention computation is now linear w.r.t. the feature map size. Moreover, the linear complexity enables a more fine-grained tuning capability to gradually increase the feature resolution while maintaining computational feasibility. Leveraging these higher-resolution features can effectively enhance the depth estimation accuracy, as we shall see in the experiments.

\textbf{Utilizing Temporal Information:} Existing works on self-supervised multi-camera depth estimation do not leverage temporal information at inference time. One key reason is the prohibitive computation and memory requirements that would be incurred when more frames were included.

Our efficient guided attention provides a viable option to incorporate more views/frames. More specifically, for each camera view, we can jointly stack features from previous frames as well as those from neighboring views, for computing keys and values in the attention. Suppose the features of the neighboring views at the current time $t$ are $F^t_{i-1,s},F^t_{i+1,s},...$ and features of the previous frames are $F^{t-1}_{i,s},F^{t-2}_{i,s},...$. The new stacked reference features for view $i$ at time $t$ is then given by $H^t_{i,s} = \text{concat}(F^t_{i-1,s},F^t_{i+1,s},...,F^{t-1}_{i,s},F^{t-2}_{i,s},...)\in \mathbb{R}^{((n_i+n_t) \cdot n_s)\times c}$, where $n_t$ is the number of previous frames to be included. Given $F^t_{i,s}$ and $H^t_{i,s}$, our guided attention can be readily computed following the steps outlined in Section~\ref{subsec:EGA}. We note that when including the features of more previous frames in the attention, the complexity of computing the attention map increases linearly. In our experiments, we include the frames from the previous timestep, which has more overlapped regions with the current frames, taking into account the fact that future frames are not available at test time for real-world applications. 

\subsection{Loss Function}
\vspace{-3pt}
We follow standard practices \cite{godard2019digging,wei2022surrounddepth} and train our entire system by minimizing the photometric error averaged over all camera views:
\begin{equation}
    \mathcal{L}_p\! =\! \underset{t^{'}}{\min}\,\frac{\alpha}{2}(1 - \text{SSIM}(I_t, \hat{I}_{t^{'}})) + (1 - \alpha)||I_t - \hat{I}_{t^{'}}||_1,
\end{equation}
where $\alpha=0.85$, $t^{'}\in\{t\!-\!1, t\!+\!1\}$ which are the two source frames from the previous and the next time steps, $\hat{I}_{t^{'}}$ is the synthesized target view, and $||\cdot||_1$ is the $L_1$ norm and $\text{SSIM}(\cdot)$ is the structural similarity measure~\cite{wang2004image}. 

An edge-aware smoothness loss: 
\begin{equation}
\mathcal{L}_s = |\partial_xd_t^{*}|e^{-|\partial_xI_t|} + |\partial_yd_t^{*}|e^{-|\partial_yI_t|}
\end{equation}
is also applied to prevent estimated depth from shrinking. 

The final training loss is then given by 
\begin{equation}
    \mathcal{L} = \mathcal{L}_p + \lambda\mathcal{L}_s
\end{equation}
where $\lambda=0.001$ is used to balance the two loss terms. 

We predict each camera pose independently instead of jointly in \cite{wei2022surrounddepth}. Also we do not perform structure-from-motion pretraining proposed in \cite{wei2022surrounddepth} as we do not attempt to recover metric depth directly. More details on training (e.g., differentiable view synthesis) can be found in the supplementary material.

\section{Experiments}
\vspace{-3pt}
We carry out extensive experiments to evaluate our proposed EGA-Depth on public large-scale benchmarks and compare with existing state-of-the-art solutions. We also conduct detailed ablation studies to provide insights into different aspects of our proposed approach.

\begin{table*}
    \small
  \centering
  \begin{tabular}{lccccccc}
    \toprule
    \textbf{Method} & \cellcolor{red!25}Abs Rel $\downarrow$ & \cellcolor{red!25}Sq Rel $\downarrow$ & \cellcolor{red!25}RMSE $\downarrow$ & \cellcolor{red!25}RMSE log $\downarrow$ & \cellcolor{blue!25}$\delta < 1.25$ $\uparrow$ & \cellcolor{blue!25}$\delta < 1.25^2$ $\uparrow$ & \cellcolor{blue!25}$\delta < 1.25^3$ $\uparrow$ \\
    \midrule
    Monodepth2~\cite{godard2019digging} & 0.287 & 3.349 & 7.184 & 0.345 & 0.641 & 0.845 & 0.925\\
    PackNet-SfM~\cite{packnet} &0.309 &2.891 &7.994 &0.390 &0.547 &0.796 &0.899\\
    FSM~\cite{guizilini2022full}& 0.299 & - & - & - & - & - & - \\
    FSM$^{*}$~\cite{guizilini2022full} & 0.334 & 2.845 & 7.786 & 0.406 & 0.508 & 0.761 & 0.894 \\
    SurroundDepth~\cite{wei2022surrounddepth} & 0.245 & 3.067 & 6.835 & 0.321 & 0.719 & 0.878 & 0.935 \\
    \hdashline
    EGA-Depth-LR (ours) & {0.239} & {2.357} & {6.801} & {0.317} & {0.723} & {0.880} & {0.936}\\
    EGA-Depth-MR (ours) & {0.228} & {2.113} & {6.738} & {0.311} & {0.728} & {0.885} & {0.940}\\
    EGA-Depth-HR (ours) & \textbf{0.223} & \textbf{1.987} & \textbf{6.599} & \textbf{0.306} & \textbf{0.732} & \textbf{0.889} & \textbf{0.942}\\
    \bottomrule
  \end{tabular}
  \vspace{-8pt}
  \caption{Quantitative evaluation of self-supervised multi-camera depth estimation on nuScenes. The best results are highlighted in bold. The row of FSM* shows the results of FSM reproduced by~\cite{wei2022surrounddepth}. Note that all the methods use the same input image resolution. LR, MR, and HR refer to the choices of internal feature resolution in our EGA-Depth approach.}
  \label{tab:nuscenes}
\end{table*}

\begin{figure*}[t!]
  \centering
   \includegraphics[width=0.99\textwidth]{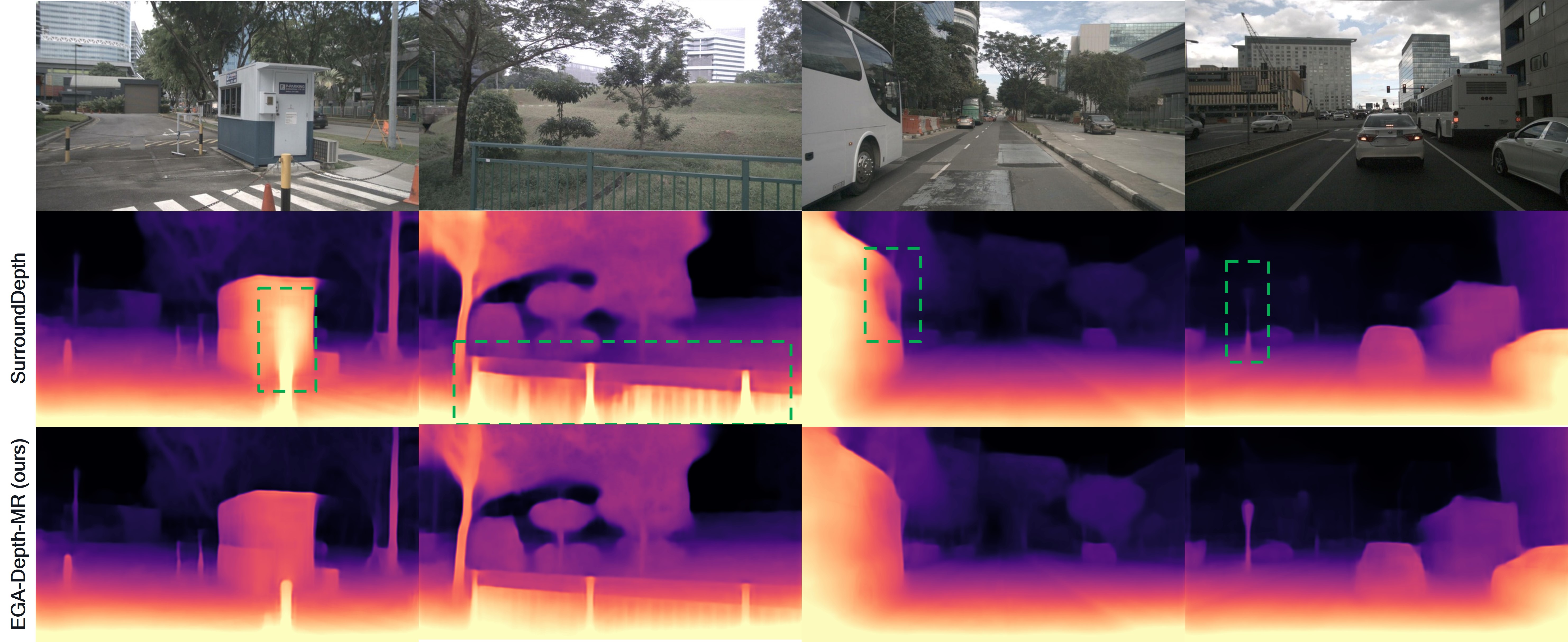}
   \vspace{-5pt}
   \caption{Qualitative comparison between EGA-Depth and SurroundDepth~\cite{wei2022surrounddepth} on nuScenes.
   The second row shows the estimated depth maps from SurroundDepth and the third row shows the estimation from our EGA-Depth method. It can be seen that EGA-Depth provides more accurate depth estimation, with better spatial consistency and finer details. The green boxes highlight sample regions where our method considerably improves the estimation quality.}
   \label{fig:comparison}
   \vspace{-5pt}
\end{figure*}

\subsection{Experimental Setup}
\vspace{-3pt}
\textbf{Datasets:} We train and evaluate EGA-Depth on two of the most popular yet challenging multi-camera autonomous driving benchmark datasets, \textit{i.e.}, nuScenes~\cite{caesar2020nuscenes} and DDAD~\cite{packnet}.

nuScenes consists of 1.4 million images from 1,000 scenes of urban driving collected in Boston and Singapore using a synchronized camera array of 6 cameras.
It is a very challenging dataset for self-supervised depth estimation since many scenes have low illumination, adverse weather conditions like rain and fog, and cluttered surroundings like construction sites. The small view overlaps across cameras further incur extra challenges for predicting consistent multi-camera depths. Following~\cite{wei2022surrounddepth}, we downsample the images from the original resolution of $900\times 1600$ to $352\times 640$. We filter out the static frames, resulting in 20,096 samples used for training and 6,019 samples for evaluation.

DDAD is an autonomous driving dataset collected in the U.S. and Japan using a synchronized 6-camera array, featuring long-range (up to 250m) and diverse urban driving scenarios. Following~\cite{wei2022surrounddepth}, we downsample the images from the original resolution of $1216\times 1936$ to $384\times 640$, resulting in 12,650 training samples and 3,950 validation samples. Following~\cite{wei2022surrounddepth, guizilini2022full}, we use the occlusion masks to reweight the photometric loss in training.

\textbf{Architecture:} We use ResNet34~\cite{he2016deep} with ImageNet~\cite{deng2009imagenet} pretrained weights as the image encoder. For each camera view, we use the two neighboring camera views when computing our guided attention. We use $Z=8$ attention heads. We consider 3 resolution options for the features to be fed into the attention, low resolution (LR), medium resolution (MR), and high resolution (HR). On nuScenes, for LR, we use $11\times20$ feature maps for all 5 scales. For MR, we use $22\times40$ for the 4 larger scales and $11\times20$ for the smallest scale. For HR, we further increase the feature map to $44\times80$ for the largest scale. For LR, we do not project the keys and values to a lower dimension since the input length is already small. For MR, we use $k_s=880$ as the fixed embedding dimension in the attention for all scales. For HR, we use $k_s=1024$ for the largest scale and $k_s=880$ for the rest. on DDAD, we use $12\times40$ feature maps for all scales for LR and $24\times40$ for all scales for MR, with no dimension reduction for LR and $k_s=960$ for MR.

\textbf{Training:} We follow the self-supervised scheme from~\cite{wei2022surrounddepth} to train our network, including learning rate scheduling and data augmentation. For the pose network, we use the ResNet18-based model from~\cite{godard2019digging}. We note that for pose estimation, we follow the standard practice of predicting each camera pose separately, instead of performing joint pose estimation which, based on our experiments, would give inferior results. All our experiments are conducted using 4 Nvidia A100 GPUs, each having 80GB of high-bandwidth memory.

\textbf{Evaluation:} We use the standard metrics to evaluate depth estimation quality~\cite{eigen2014depth, godard2019digging}, including Absolute Relative Error (Abs Rel), Squared Relative Error (Sq Rel), Root-Mean-Squared-Error (RMSE), RMSE log, $\delta < 1.25$, $\delta < 1.25^2$, and $\delta < 1.25^3$.\footnote{We refer readers to the supplementary file for detailed mathematical definitions of these metrics.} Following previous works~\cite{guizilini2022full, wei2022surrounddepth}, we apply median-scaling at test time and report numbers averaged over all cameras. The depth values are evaluated with maximum distances of 80m and 200m, for nuScenes and DDAD, respectively. 

\subsection{Evaluation Results}
\label{eval-results}
\vspace{-3pt}
We report both quantitative and qualitative results on nuScenes and DDAD, and compare with baseline and latest methods on self-supervised multi-camera depth estimation, including Monodepth2~\cite{godard2019digging} and PackNet-SfM~\cite{packnet} (treating each view independently), FSM~\cite{guizilini2022full}, and SurroundDepth~\cite{wei2022surrounddepth}.

\textbf{nuScenes:}
The performance evaluation on nuScenes is shown in Table~\ref{tab:nuscenes} and Figures~\ref{fig:nuscenesres}~and~\ref{fig:comparison}. Results shown in Figs.~\ref{fig:nuscenesres}~and~\ref{fig:comparison} are generated using the EGA-Depth-MR. In Table~\ref{tab:nuscenes}, it can be seen that even our lightest model, EGA-Depth-LR, which uses low-resolution feature maps, already outperforms the latest SOTA across all metrics by a considerable margin. By leveraging higher resolutions for the features, we can further reduce depth estimation errors. Specifically, EGA-Depth-HR reduces the squared relative error by $35\%$ when compared to the latest SOTA method of SurroundDepth.

Fig.~\ref{fig:nuscenesres} shows the multi-camera depth estimation by our proposed EGA-Depth on two sample scenes. It can be seen that EGA-Depth accurately predicts the depths (e.g., around the vehicles), even under adverse weather conditions like rain. In Fig.~\ref{fig:comparison}, we compare the estimated depth maps between our EGA-Depth and SurroundDepth. We see that EGA-Depth can predict depths with sharper details and more robustness, thanks to using higher-resolution feature maps ($4\times$ larger than in SurroundDepth)

\begin{figure*}[t!]
  \centering
   \includegraphics[width=0.99\textwidth]{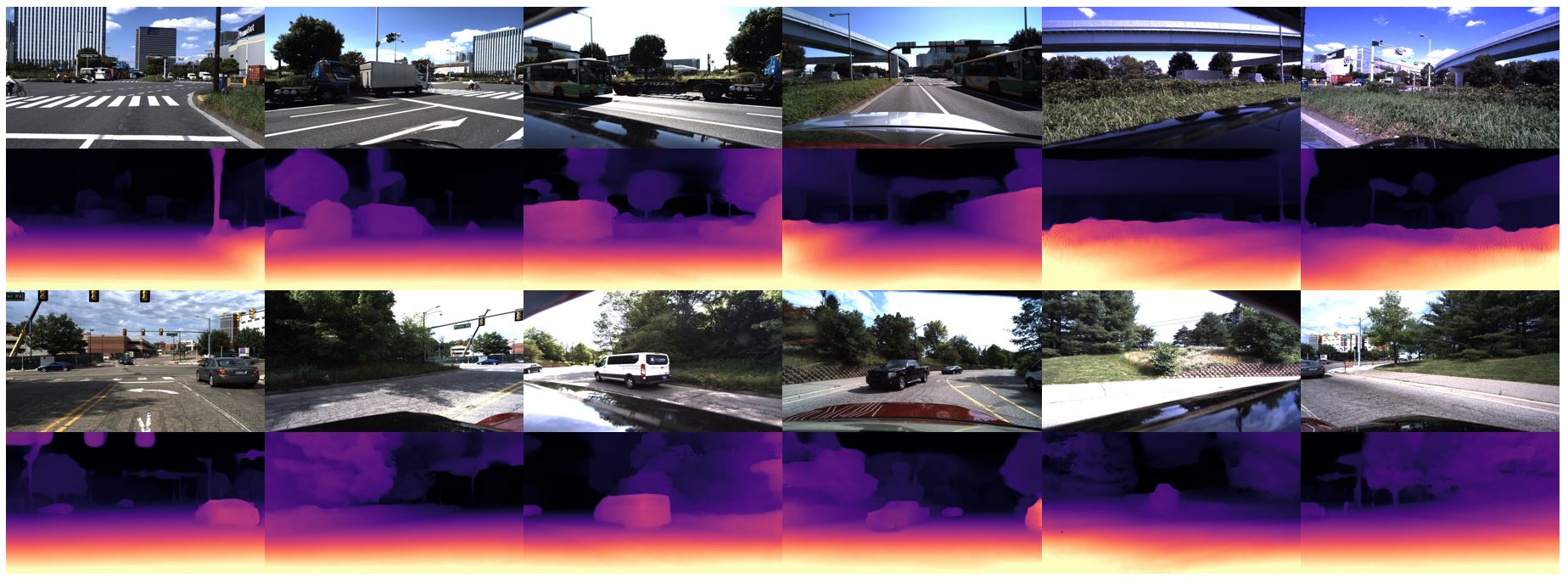}
   \vspace{-8pt}
   \caption{Qualitative results of self-supervised multi-camera depth predictions on two sample scenes from the DDAD dataset.
   For each scene, we show the front, front-left, back-left, back, back-right, and front-right camera views from left to right. We can see that our EGA-Depth can generate accurate depth maps with fine details. }
   \label{fig:ddadres}
\end{figure*}

\begin{table*}
\small
  \centering
  \begin{tabular}{lccccccc}
    \toprule
    \textbf{Method} & \cellcolor{red!25}Abs Rel $\downarrow$ & \cellcolor{red!25}Sq Rel $\downarrow$ & \cellcolor{red!25}RMSE $\downarrow$ & \cellcolor{red!25}RMSE log $\downarrow$ & \cellcolor{blue!25}$\delta < 1.25$ $\uparrow$ & \cellcolor{blue!25}$\delta < 1.25^2$ $\uparrow$ & \cellcolor{blue!25}$\delta < 1.25^3$ $\uparrow$ \\
    \midrule
    Monodepth2 \cite{godard2019digging} & 0.217 & 3.641 & 12.962 & 0.323 & 0.699 & 0.877 & 0.939\\
    PackNet-SfM~\cite{packnet} &0.234 &3.802 &13.253 &0.331 &0.672 &0.860 &0.931\\
    FSM \cite{guizilini2022full} & 0.202 & - & - & - & - & - & - \\
    FSM$^{*}$ \cite{guizilini2022full} & 0.229 & 4.589 & 13.520 & 0.327 & 0.677 & 0.867 & 0.936 \\
    SurroundDepth \cite{wei2022surrounddepth} & 0.200 & 3.392 & 12.270 & 0.301 & 0.740 & 0.894 & 0.947 \\
    \hdashline
    EGA-Depth-LR (ours) & {0.195} & {3.211} & {12.117} & {0.297} & {0.743} & {0.896} & 0.947\\
    EGA-Depth-MR (ours) & \textbf{0.191} & \textbf{3.126} & \textbf{11.922} & \textbf{0.290} & \textbf{0.747} & \textbf{0.901} & \textbf{0.950}\\
    \bottomrule
  \end{tabular}
  \vspace{-8pt}
  \caption{Quantitative evaluation of self-supervised multi-camera depth estimation on DDAD. The best results are highlighted in bold. The row of FSM* shows the results of FSM reproduced by~\cite{wei2022surrounddepth}. Note that all the methods use the same input image resolution. LR, MR, and HR refer to the choices of internal feature resolution in our EGA-Depth approach.}
  \label{tab:ddad}
  \vspace{-8pt}
\end{table*}

\textbf{DDAD:} The quantitative evaluation results on DDAD are shown in Table~\ref{tab:ddad}. We see that our EGA-Depth models consistently outperform the existing SOTA methods across all the metrics. Fig.~\ref{fig:ddadres} shows visual examples of predicted depth maps by using EGA-Depth-MR. It can be seen that our predicted depth maps are very accurate and preserve important details, such as object boundaries.

\begin{table}[t!]
\vspace{-0pt}
\small
  \centering
  \begin{tabular}{lccc}
    \toprule
    \textbf{Method} & \textbf{Pre. Frame} & \cellcolor{red!25}$\!$Abs Rel $\downarrow\!$ & \cellcolor{blue!25}$\!\delta\!<\!1.25$ $\uparrow\!$ \\
    \midrule
    SurroundDepth\cite{wei2022surrounddepth} &  & 0.245 & 0.719\\
    SurroundDepth-T & \checkmark & 0.368 & 0.501\\
    \midrule
    EGA-Depth-LR    &  & {0.239} &{0.723}\\
    EGA-Depth-LR-T  & \checkmark & {0.237} &{0.723}\\
    EGA-Depth-MR    & &{0.228} &{0.728}\\
    EGA-Depth-MR-T  &\checkmark &{0.226} &{0.729}\\
    \bottomrule
  \end{tabular}
  \vspace{-8pt}
  \caption{Results of leveraging features from the previous time step on nuScenes.}
  \label{tab:temporal}
  \vspace{-10pt}
\end{table}

\textbf{Using Previous Frame:} As described in Section~\ref{subsec:high_res_temporal}, EGA-Depth can take advantage of previous frames in our guided attention module. 
It can be seen in Table~\ref{tab:temporal}, adding frames from the previous timestep consistently improves the depth estimation accuracy of EGA-Depth. 
We note that the accuracy improvements are not very significant. This is due to the fact that the curated nuScenes video sequences (containing only key frames) have a low frame rate of 2Hz and hence, view overlaps between consecutive frames are small. As a result, the amount of temporal information that can be leveraged is limited by the data. 
\begin{table*}[t!]
\small
  \centering
  \begin{tabular}{lcccccc}
    \toprule
    \textbf{Attention Choice} & \textbf{Feature Map Resolution} & $k_s$ & $n_i\cdot n_s$ & \cellcolor{red!25}Abs Rel $\downarrow$& \cellcolor{blue!25}$\delta < 1.25$ $\uparrow$ & \cellcolor{red!25}GFLOPs $\downarrow$ \\
    \midrule
    SurroundDepth~\cite{wei2022surrounddepth} & $\{11\times 20\}_{\times 5}$ & - & 1320 & 0.245 & 0.719 & 132.32\\
    \midrule
    SurroundDepth-MR & \makecell{$\{22\times 40\}_{\times 4}$\\$\{11\times 20\}_{\times 1}$} & \makecell{880\\-} & \makecell{5280\\1320} & 0.236 &  0.721 & 424.01\\
    \midrule
    EGA-Depth-LR & $\{11\times 20\}_{\times 5}$ & - & 440 & {0.239} & {0.723} & {64.94}\\
    \midrule
    
    EGA-Depth-MR & \makecell{$\{22\times 40\}_{\times 4}$\\$\{11\times 20\}_{\times 1}$} & \makecell{880\\-} & \makecell{1760\\440} & {0.228} & {0.728} & 214.81\\
    \midrule
    EGA-Depth-HR & \makecell{$\{44\times 80\}_{\times 1}$\\$\{22\times 40\}_{\times 3}$\\$\{11\times 20\}_{\times 1}$} & \makecell{1024\\880\\-} &  \makecell{7040\\1760\\440} & {0.223} & {0.732} & 475.79\\
    \bottomrule
  \end{tabular}
  \vspace{-8pt}
  \caption{Results of computing attention over feature maps of different resolutions on nuScenes. 
  The first column shows the two attention choices of SurroundDepth and our EGA-Depth. The second column shows the resolutions of the 5-scale (from top to bottom) feature maps going into the attention module. 
  $k_s$ is the projection dimension that is used to map the length of the flattened and concatenated neighboring-view features, $n_i\cdot n_s$, to a constant value. SurroundDepth-MR is our extension of SurroundDepth using higher feature resolutions; note that it is infeasible to scale up the feature maps of the original SurroundDepth, even when using 4 Nvidia A100 GPUs. On the other hand, EGA-Depth can process features with a resolution as large as $44\times 80$ for $352\times 640$ input images.}
  \label{tab:abl1}
  \vspace{-5pt}
\end{table*}
As a baseline for our temporal studies, we augment our SurroundDepth implementation to take advantage of previous frames. More specifically, we feed the features of 12 views/frames (6 for the current time step and 6 for the previous time step) to its self-attention module. We note that this is consistent with the original SurroundDepth design. While this almost doubled the amount GFLOPs needed (going from 132.32 to 220.15), the accuracy becomes worse. This can be attributed to model overfitting, since much more parameters are used in the model, and attention computation over little-/non-overlapping views, causing the model to potentially learn spurious correlations. On the other hand, when using the previous frames, EGA-Depth-LR-T only incurs a small increase in GFLOPs (going from 64.94 to 91.56) while also clearly improving the depth estimation accuracy.
\subsection{Ablation Studies}
\label{ablations} \vspace{-3pt}

\vspace{0pt}
\noindent\textbf{Feature Map Resolution:} 
Table~\ref{tab:abl1} shows details of the feature map resolution and projection dimension ($k_s$) choices of our EGA-Depth models, as well as their accuracies and computational costs. 
When using low-resolution features in EGA-Depth-LR, we do not reduce the key and value dimensions since the input length (i.e., $H_s \times W_s$ of the feature map) is already small with $n_s=440$. EGA-Depth-LR already outperforms SurroundDepth and uses $51\%$ less computation.
Our EGA-Depth-MR and EGA-Depth-HR models employ higher resolutions for the feature maps, leading to improved depth estimation accuracy with increased computational costs.

In order to investigate the performance of SurroundDepth with higher resolution features, we implement a baseline, SurroundDepth-MR. In order to make the computation feasible, we replace the standard self-attention with Linformer and increase the feature map resolution by $4\times$ except for the smallest scale. Although SurroundDepth-MR has better accuracy than the original SurroundDepth, it requires significantly more computations. In particular, SurroundDepth-MR can only achieve slightly better accuracy than EGA-Depth-LR, but requires $4.5\times$ GFLOPs.

Our proposed EGA-Depth provides much more favorable accuracy-efficiency trade-offs as compared to using the SurroundDepth attention scheme, as also shown in Figure~\ref{fig:accuracy-efficiency-tradeoff}. Moreover, even when using 4 Nvidia A100 GPUs, further scaling up the feature maps of SurroundDepth (using linear transformer) results in infeasible memory costs.



\vspace{2pt}
\noindent\textbf{Projection Dimension $k_s$:} 
In Table~\ref{tab:abl3}, we show how the choice of the projection dimension $k_s$ impacts the final depth estimation performance. For EGA-Depth-MR, we set $k_s=880$, which reduces the query and key dimensions by half. We also experiment with other values of $k_s$. It can be seen that all these $k_s$ values allow our model to generate more accurate depth estimation as compared to SurroundDepth. However, we observe that increasing $k_s$ does not necessarily map to monotonically improving results.
\begin{table}
\small
  \centering
  \begin{tabular}{lcccc}
    \toprule
    \textbf{Method} & $\!k_s\!$ & $\!n_i\! \cdot \!n_s\!$ & \cellcolor{red!25}$\!$Abs Rel $\downarrow\!$ & \cellcolor{blue!25}$\!\delta\! <\! 1.25$ $\uparrow\!$ \\
    \midrule
    SurroundDepth~\cite{wei2022surrounddepth} & - & 1320 & 0.245 & 0.719\\
    \hline
    EGA-Depth-MR & \makecell{512\\880\\1024\\1536} & \makecell{1760} &\makecell{0.233\\{0.228}\\0.229\\0.235\\}& \makecell{0.721\\{0.728}\\0.728\\0.723\\}\\
    \bottomrule
  \end{tabular}
  \vspace{-8pt}
  \caption{Results on nuScenes of using different values of $k_s$ when projecting keys and values to a lower dimension.}
  \label{tab:abl3}
  \vspace{-10pt}
\end{table}

\section{Conclusion}
\vspace{-3pt}
In this paper, we presented a novel, efficient guided attention approach, EGA-Depth, for self-supervised multi-camera depth estimation. In contrast to existing SOTA that applied standard self-attention to all the camera views, we proposed to utilize attention across each camera view and its neighboring views, which significantly reduces computational costs while improving accuracy. We further incorporated an operation to project the keys and values of the attention to lower, prescribed dimensions, which removed the quadratic complexity with respect to the input feature map resolution. Based on our inherently efficient design, we leveraged higher feature resolutions in order to further improve our depth estimation accuracy. Our EGA-Depth framework can also readily incorporate frames from previous time steps for attention. As we have shown in our experiments on challenging benchmark datasets of nuScenes and DDAD, our proposed EGA-Depth sets the new SOTA accuracy while our design points lie on the Pareto frontier of accuracy-efficiency trade-off.

\newpage
{\small
\bibliographystyle{ieee_fullname}
\bibliography{egbib}
}

\end{document}